\documentclass[times,twocolumn,final]{elsarticle}

\usepackage{arxiv}
\usepackage{graphicx}%
\usepackage{multirow}%
\usepackage{amsmath,amssymb,amsfonts}%
\usepackage{amsthm}%
\usepackage{mathrsfs}%
\usepackage[title]{appendix}%
\usepackage{xcolor}%
\usepackage{booktabs}%
\usepackage{algorithm}%
\usepackage{algorithmicx}%
\usepackage{algpseudocode}%
\usepackage{pifont}
\newcommand{\cmark}{\ding{51}}%
\newcommand{\xmark}{\ding{55}}%
\usepackage{adjustbox}
\usepackage{graphicx}
\usepackage{hyperref}
\usepackage{float}
\usepackage{placeins}
\usepackage{fancyhdr}

\fancypagestyle{firstpagestyle}{
    \fancyhf{} 
    \fancyhead{} 
    \fancyfoot{} 
    \fancyhead[CO]{\em \fontsize{9pt}{8pt}\selectfont This article has been accepted to the International Conference on Information Processing in Computer-Assisted Interventions, 2024.}
}

\fancypagestyle{default}{
    \fancyhf{}
    \fancyhead[R]{\thepage} 
    \fancyhead[LO]{\em \fontsize{9pt}{8pt}\selectfont This article has been accepted to the International Conference on Information Processing in Computer-Assisted Interventions, 2024.}
}

\begin{document}
\title{Optimizing Latent Graph Representations of Surgical Scenes for Zero-Shot Domain Transfer}

\author[1]{Siddhant \snm{Satyanaik}
\corref{contrib}}
\author[1]{Aditya \snm{Murali}\corref{contrib}\fnref{corresp}}
\cortext[contrib]{These authors contributed equally to this work.}
\fntext[corresp]{Corresponding author: \texttt{murali@unistra.fr}}
\author[1]{Deepak \snm{Alapatt}}
\author[4]{Xin \snm{Wang}}
\author[2,3]{Pietro \snm{Mascagni}}
\author[1,2]{Nicolas \snm{Padoy}}

\address[1]{ICube, University of Strasbourg, CNRS, France}
\address[2]{IHU Strasbourg, France}
\address[3]{Fondazione Policlinico Universitario A. Gemelli IRCCS, Rome, Italy}
\address[4]{West China Hospital of Sichuan University, Chengdu, China}

\received{XXX}
\finalform{XXX}
\accepted{XXX}
\availableonline{XXX}
\communicated{XXX}

\begin{abstract}
\textbf{Purpose}:
Advances in deep learning have resulted in effective models for surgical video analysis; however, these models often fail to generalize across medical centers due to domain shift caused by variations in surgical workflow, camera setups, and patient demographics. Recently, object-centric learning has emerged as a promising approach for improved surgical scene understanding, capturing and disentangling visual and semantic properties of surgical tools and anatomy to improve downstream task performance.
In this work, we {conduct a multi-centric performance benchmark of object-centric approaches, focusing on Critical View of Safety assessment in laparoscopic cholecystectomy, then propose an improved approach for unseen domain generalization.}

\noindent\textbf{Methods}:
We evaluate four object-centric approaches {for domain generalization}, establishing baseline performance. {Next, leveraging the disentangled nature of object-centric representations, we dissect one of these methods through a series of ablations (e.g. ignoring either visual or semantic features for downstream classification).
Finally, based on the results of these ablations, we develop an optimized method specifically tailored for domain generalization, LG-DG, that includes a novel
disentanglement loss function.}

\noindent\textbf{Results}:
Our optimized approach, LG-DG, achieves an improvement of {9.28\% over the best baseline approach}. More broadly, we {show that} object-centric approaches are highly effective for domain generalization {thanks to} their modular approach to representation learning.

\noindent\textbf{Conclusion}:
We investigate {the use of object-centric methods for unseen domain generalization, identify method-agnostic factors critical for performance, and present an optimized approach that substantially outperforms existing methods.}
\\

\end{abstract}

\maketitle
\thispagestyle{firstpagestyle}

\begin{figure*}[h!]
    \centering
    \includegraphics[width=\linewidth]{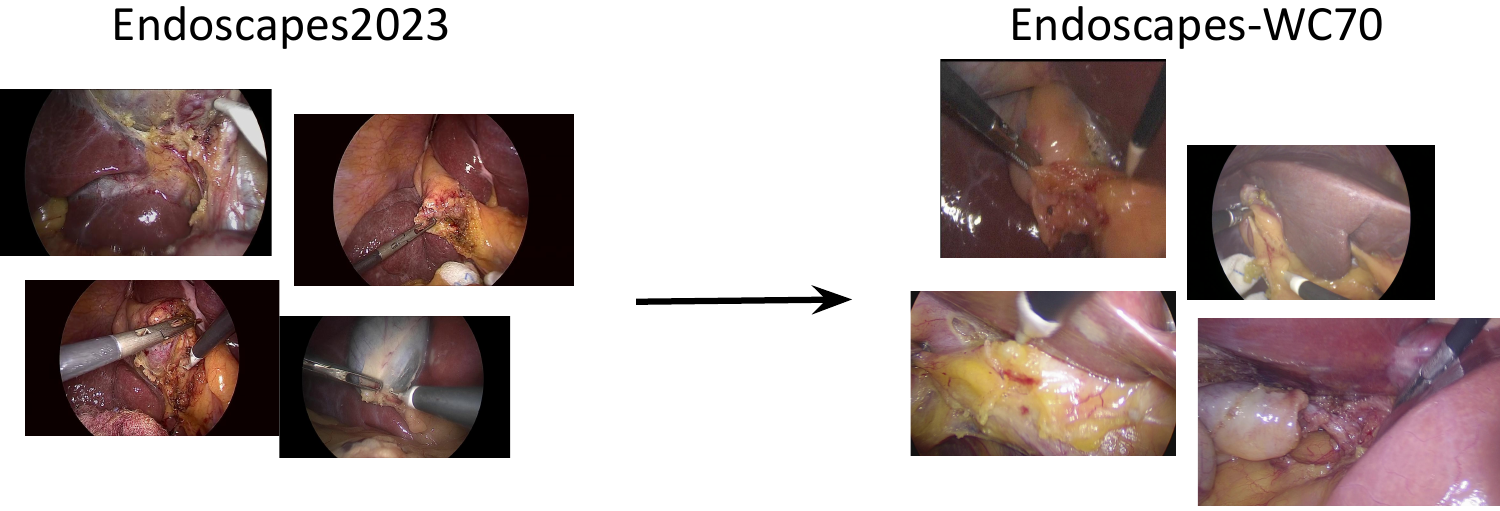}
    \caption{Qualitative examples to illustrate the visual domain gap between Endoscapes2023 and Endoscapes-WC70. The images differ in aspect ratio, color distribution, and field of view that could be caused by variations in the laparoscope used and in surgical workflow.}
    \label{fig:object_detector}
\end{figure*}

\section{Introduction}\label{sec1}

Surgical video analysis is a rapidly growing field that aims to mine critical information from unstructured surgical videos at scale, which can then be used to enhance various aspects of surgical practice.
Over the past decade, many works have harnessed advances in deep learning to achieve proficiency in tasks like automated phase recognition~\cite{twinanda2016endonet}, tool and anatomy segmentation~\cite{grammatikopoulou2021cadis,sestini2023fun}, and fine-grained action recognition~\cite{sharma2023surgical,hao2023act}.
Yet, practical deployment of these models requires effective \textit{domain generalization}, or in other words, generalization to distinct hospital environments characterized by variations in surgical workflow, instrumentation, camera setups, and patient demographics.

A straightforward {approach to this problem} is to {simply} build training datasets comprising annotated videos from numerous surgical centers; however, this is often impractical due to multifaceted difficulties and privacy concerns regarding data collection and sharing~\cite{kassem2022federated}.
Moreover, even if these concerns can be overcome, annotating these datasets and ensuring consistency across centers is a critical bottleneck, particularly for fine-grained tasks like semantic segmentation and action recognition.
As a result, many works in medical and surgical computer vision have approached multicentric generalization as a domain adaptation problem, aiming to adapt models trained for one center (source domain) to other centers (target domains)~\cite{srivastav2021adaptor,wang2019unifying,mottaghi2022adaptation,xu2022deep}.
In this context, the diverse aforementioned data and annotation availability scenarios can be broadly divided into three different problem settings: (1) Fully Supervised/Supervised Domain Adaptation (SDA), where data and annotations are available for the source and target domain, (2) Unsupervised Domain Adaptation (UDA), where data from the target domain is available but not annotations, and (3) {Domain Generalization (DG)}, where no information about the target domain is known.

While several works have focused on UDA, the {DG} setting is relatively under-explored; nevertheless, it is especially relevant in the surgical domain, where collecting and sharing data is still extremely challenging.
Consequently, in this work, we focus on the latter, specifically aiming to improve performance by leveraging \textit{object-centric classification} approaches.
Object-centric methods learn image representations that are strongly conditioned on the objects in the scene, often by including an object detection step prior to classification.
These representations can then be finetuned for different downstream tasks, enabling improved performance.
Our key observation is that such methods could be {highly} effective for domain generalization because they explicitly enforce downstream predictions to be conditioned on detected objects rather than extraneous factors like lighting or background artifacts.

To explore this hypothesis, we begin by thoroughly benchmarking domain adaptation performance of four different object-centric methods, starting with the fully supervised and SDA settings to establish ceiling performances, then moving to the DG setting.
We focus on Critical View of Safety (CVS) prediction, { for two reasons: (1) it is a challenging fine-grained task} that has been {successfully} tackled using object-centric methods~\cite{mascagni2021artificial,murali2023latent,murali2023encoding}, {and (2) exploration of multicentric generalization in the context of such fine-grained tasks is limited, raising the need for further investigation.}
To conduct a multicentric analysis, we employ {the Endoscapes2023 dataset~\cite{murali2023endoscapes}}, collected in Strasbourg, France, and additionally introduce \textbf{Endoscapes-WC70}, a dataset of 70 laparoscopic cholecystectomy videos collected in Sichuan, China annotated with CVS criteria and segmentation masks.
Finally, we {propose an object-centric approach for improved domain generalization, \textbf{LG-D}omain\textbf{G}en \textbf{(LG-DG)}, that extends the best single-center method}, LG-CVS~\cite{murali2023latent}, {with a disentanglement loss function to learn more robust representations and thereby improve} domain generalization.

In our experiments, we find that object-centric models generally outperform non-object-centric classifiers for domain generalization{, and pinpoint critical attributes across methodologies for effective domain generalization.
Our proposed approach incorporates these various principles into a single method, thereby outperforming existing approaches.}

Our contributions can be summarized as follows:
\begin{enumerate}
    \item We study 4 object-centric methods in the context of domain generalization, focusing on Critical View of Safety prediction.
    \item We propose an improved latent graph-based object-centric approach{, LG-DG, that substantially outperforms existing approaches for domain generalization.}
    \item We introduce the Endoscapes-WC70 dataset, which comprises 7,690 images from 70 videos annotated with CVS criteria, of which 510 are additionally annotated with segmentation masks.
\end{enumerate}

\begin{figure*}[h!]
    \centering
    \includegraphics[width=\linewidth]{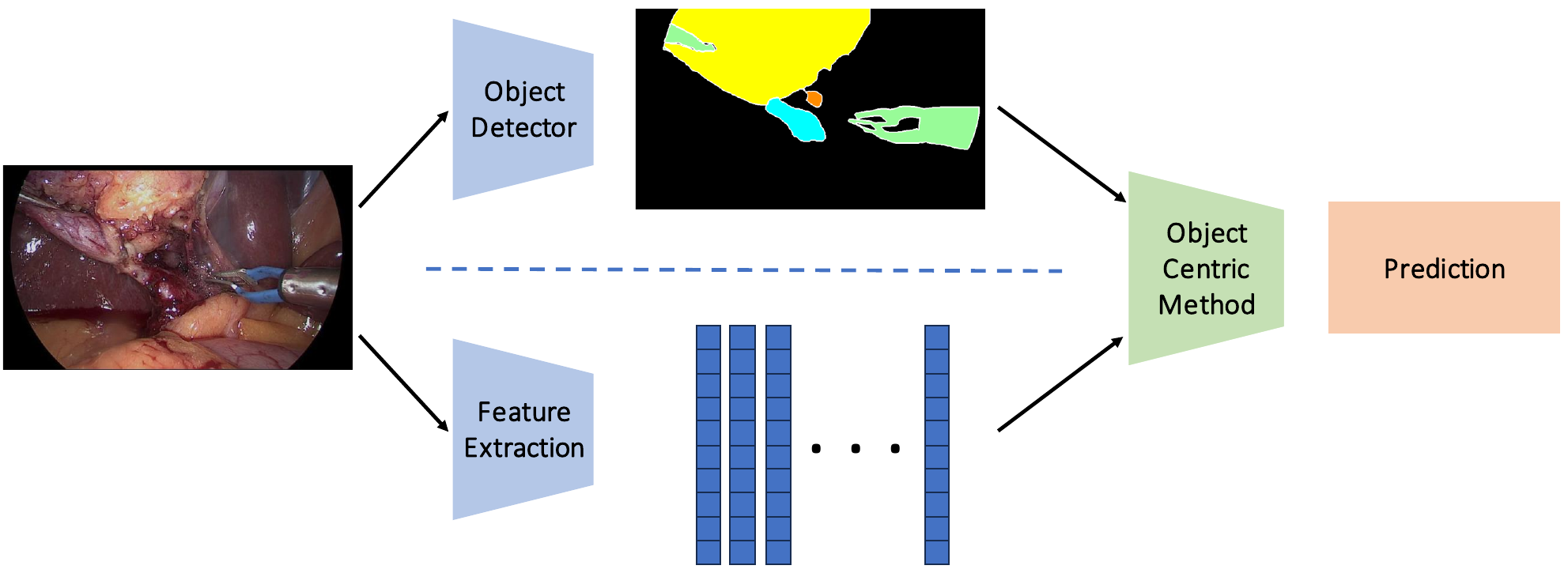}
    \caption{An illustration of a generic Object-Centric method.}
    \label{fig:object_detector}
\end{figure*}

\thispagestyle{default}

\section{Related Work}\label{secrw}
\subsection{Domain Generalization}
{Annotating surgical data is extremely expensive due to the need for expert knowledge. To alleviate this burden, the research community has actively explored various domain adaptation methodologies, such as unsupervised domain adaptation (UDA)~\cite{srivastav2021adaptor,wang2019unifying}, where target data but not labels are used to refine source domain-trained models, and semi-supervised domain adaptation (SSDA)~\cite{mottaghi2022adaptation,basak2023semi}, where some of the target data also contains labels.}
A related research area is federated learning (FL), which restricts data sharing among domains, but allows decentralized model training using all available data.
Federated learning has been explored extensively in the medical imaging community~\citep{sohan2023systematic}, and more recently for surgical phase recognition~\citep{kassem2022federated}.

{Unlike UDA, SSDA, and FL, Domain Generalization (DG)} eliminates the need for target domain data altogether.
{DG methods aim to build a single model that can generalize to unseen target domains.
Popular approaches include model ensembling, data augmentation/generation to enhance the training set with estimations of out-of-domain data, and most commonly, representation learning-based approaches that seek to learn domain-invariant representations, generally through custom model or loss function design.~\cite{choi2021robustnet,trid}}

{
Our work focuses on the under-explored area of Domain Generalization in surgical computer vision, falling under the umbrella of Representation Learning-based DG.
Unlike previous works, we investigate the use of object-centric representations, which are already disentangled feature representations, and additionally propose an auxiliary learning objective to further boost feature disentanglement, thereby improving generalization.}

\subsection{Object-Centric Learning}
Object-centric learning focuses on learning scene representations in terms of the present objects, then using these representations for downstream tasks.
It has been gaining prominence in the surgical domain, having been applied for activity recognition~\cite{hamoud2023st,ozsoy2023labrad} and phase recognition~\cite{holm2023dynamic, murali2023encoding}, action triplet recognition~\cite{sharma2023surgical}, scene captioning~\cite{pang2022rethinking}, and most relevantly, CVS prediction~\cite{murali2023latent, murali2023encoding}.

While these works focus on solving specific downstream tasks, we are instead interested in leveraging object-centric learning for {improved domain generalization}.

\thispagestyle{default}

\section{Methods}

In this section, we begin by detailing the datasets used in our multicentric study {and describe the CVS prediction downstream task.
Then, we describe the four object-centric approaches that we investigate.
Finally, we describe our methodology for optimizing LG-CVS~\cite{murali2023latent}, a state-of-the-art object-centric method for CVS prediction, to develop our proposed approach: LG-\textbf{D}omain\textbf{G}en, or LG-DG.
This consists of preliminary ablations to understand elements of object-centric methods that aid domain generalization, followed by the development of a disentanglement loss function to boost latent graph robustness, and, as a result, domain generalization performance.}

\subsection{Datasets and Downstream Task}

{The Critical View of Safety, or CVS, is a measure to assess dissection quality and exposure of key anatomical structures; proper achievement of the CVS is strongly associated with reduced adverse outcomes, such as bile duct injury.
CVS consists of three different binary criteria, detailed in~\cite{murali2023endoscapes}, making CVS prediction a multi-label classification problem.}

{To enable a multi-centric study, we use two datasets: (1) Endoscapes2023, introduced in~\cite{murali2023endoscapes} and collected in Strasbourg, France, and (2) Endoscapes-WC70, which we introduce here, collected in Sichuan, China.
Endoscapes2023 contains 58585 frames from 201 laparoscopic cholecystectomy videos; we use two subsets of Endoscapes2023: Endoscapes-CVS201, which contains 11090 frames annotated with CVS, and Endoscapes-Seg201, which contains segmentation masks for 1933 of the previous 11090 frames.}
We adopt the {official} dataset splits, using 120 videos for training, 41 for validation, and 40 for testing.
{Meanwhile, we construct Endoscapes-WC70 by collecting 70 laparoscopic cholecystectomy videos from the West China Hospital, Sichuan University, Sichuan, China; following the protocol of Endoscapes2023~\cite{murali2023endoscapes},} we annotate CVS at 5 second intervals and segmentation masks at 30 second intervals.
{Finally, we split the 70 videos into 40 training, 15 validation, and 15 test videos, applying stratified sampling based on video-level CVS achievement as done in~\cite{murali2023latent}.}
\autoref{tab:achievement_rates} shows the per-criterion CVS achievement rates for the two datasets. Endoscapes-WC70 is {particularly} imbalanced with respect to CVS achievement; we discuss the impact of this imbalance in our results.

\begin{table}[]
\setlength\tabcolsep{3pt}
\centering
\caption{Frame-Level Achievement Rates (\%) of each CVS Criterion.}
\label{tab:achievement_rates}
\begin{tabular}{ccccccc}
\multirow{2}{*}{\textbf{Criterion}} & \multicolumn{3}{c}{\textbf{Endoscapes2023}} & \multicolumn{3}{c}{\textbf{Endoscapes-WC70}} \\
                                    & Train         & Val         & Test        & Train          & Val          & Test         \\ \hline
C1: Two Structures                  & 15.6          & 16.3        & 24.0        & 2.1            & 2.0          & 3.2          \\
C2: HCT Dissection                  & 11.2          & 12.5        & 17.1        & 7.1            & 7.3          & 11.8         \\
C3: Cystic Plate                    & 17.9          & 16.7        & 27.1        & 8.4            & 9.0          & 14.9
\end{tabular}
\end{table}

\begin{figure*}
    \centering
    \includegraphics[width=\linewidth]{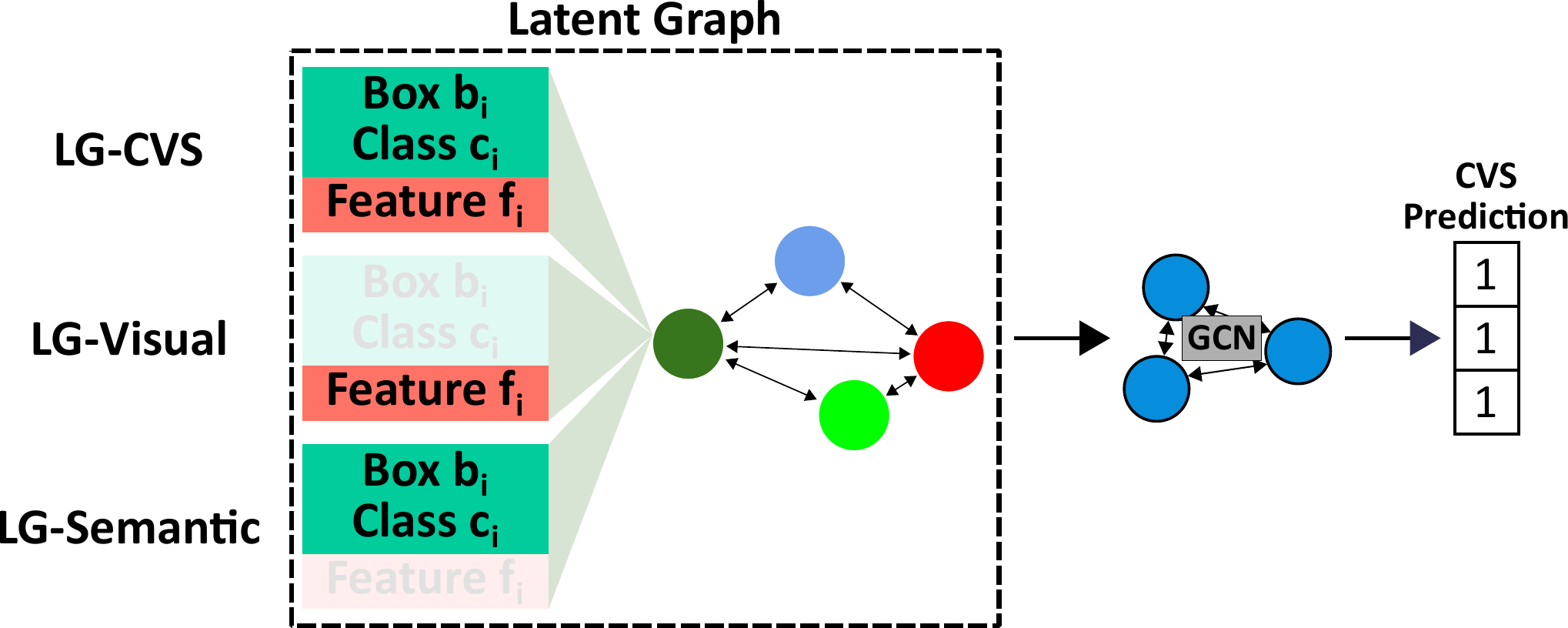}
    \caption{Three different examples of the masked latent graph $\hat{G}_{\text{CVS}}$ that is passed to the downstream classification head $\phi_{\text{CVS}}$ for CVS prediction. Each node in the graph corresponds to an object in the image. The masking operation, while pictured for a single node, is applied to all nodes.}
    \label{fig:ablations1}
\end{figure*}

\subsection{Baseline Models}
We evaluate 4 different object-centric classification methods: LG-CVS~\cite{murali2023latent}, DeepCVS~\cite{mascagni2021artificial}, LayoutCVS (an ablation of DeepCVS introduced in~\cite{murali2023latent}), and ResNet50-DetInit (also introduced in~\cite{murali2023latent}); to ensure fair comparisons, we train a single Mask-RCNN object detector to be used by all the methods.
We additionally evaluate a vanilla ResNet50 model to isolate improvements in generalization brought by object-centric modeling.
We briefly describe each approach below:\\

\noindent \textbf{ResNet50}: We finetune an ImageNet-pretrained ResNet50 classifier for CVS prediction, requiring no additional labels in the source domain. This baseline allows us to study whether object-centric approaches as a whole can enable better generalization.\\

\noindent \textbf{DeepCVS}~\cite{mascagni2021artificial}: DeepCVS is a two-stage model that consists of first segmenting an image, then concatenating the predicted segmentation mask with the original image and processing the result with a convolutional neural network to predict CVS. In doing so, the downstream CVS predictions are based on both semantic information (identity and location of various anatomical structures, encoded as a segmentation mask), and visual information (from the original image).
For fair comparisons, we use the adapted {and better performing version of} DeepCVS presented in~\cite{murali2023latent}.\\

\noindent \textbf{LayoutCVS}: LayoutCVS is identical to DeepCVS except that it does not concatenate the original image with the output of the object detector.
Including this model allows us to examine the varying robustness of semantic and visual features to shifts in input domain for downstream classification, especially through comparisons with DeepCVS.\\

\noindent \textbf{ResNet50-DetInit}: We initialize a ResNet50 classifier using trained object detection model weights, then finetune for CVS prediction, thus utilizing the fine-grained scene information encoded implicitly in the object detector's learned visual features.
{This baseline allows us to isolate the impact of visual information, but unlike the ResNet50 baseline, the visual information is sufficiently fine-grained for effective CVS prediction.}\\

\noindent \textbf{LG-CVS}: A two-stage approach for object-centric classification: in the first stage, {an encoder $\Phi_{\text{LG}}$ encodes an} image $I$ as a latent graph $G$, wherein nodes represent objects (tools and anatomical structures) and edges capture 2D geometric relationships between objects.
These nodes and edges contain object/relation-specific visual features as well as semantic features (bounding box coordinates and class probabilities).
The latent graph $G$ is then passed to a {GNN-based classification head} $\phi_{\text{CVS}}$ for downstream CVS prediction, and an auxiliary decoder $\phi_{\mathcal{R}}$ that reconstructs the original image given $G$ and a backgroundized version of $I$ (foreground regions replaced with noise).
This auxiliary reconstruction objective helps learn more discriminative object features, improving downstream performance.

\thispagestyle{default}

\subsection{{Investigating Feature Disentanglement in the Latent Graph}}
A key facet of LG-CVS is its disentanglement of object-level semantic and visual properties, encoded in the latent graph nodes, and image-level visual properties, captured by the backbone feature map.
While these properties contribute synergistically in the single-center setting~\cite{murali2023latent}, {we note that} this may not be the case for domain generalization, where errors in individual modes of information can derail model predictions.

To study this effect, we conduct a series of ablations where we mask one or more information categories from the latent graph {for both training and testing before passing to the downstream classification head $\phi_{\text{CVS}}$ and the auxiliary reconstruction head $\phi_{\mathcal{R}}$.}
\autoref{fig:ablations1} illustrates {the most extreme ablation settings}, while \autoref{tab:cvs_head_ablations} lists the various latent graph configurations that we evaluate.

{
\noindent \textbf{CVS Head Ablations.} In this first study, we experiment with different masking combinations, but only apply the masking before the downstream head, leaving the input to the reconstruction head intact.
By doing so, we aim to identify whether (1) certain feature categories are more domain-invariant then others and (2) the different feature categories contribute synergistically when the model is exposed to domain shift.
Concretely:
\begin{equation*}
    G = \Phi_{\text{LG}}(I);\ \mathcal{M}(G, c) = \hat{G}_{\text{CVS}};\
\end{equation*}

\begin{equation}
    \hat{y} = \phi_{\text{CVS}}(\hat{G}_{\text{CVS}});\ \hat{I} = \phi_{\mathcal{R}}(G),
\end{equation}
where $\mathcal{M}$ is a masking function that replaces the original values with Gaussian noise, $c$ is the feature category to mask, $\hat{G_{\text{CVS}}}$ is the  masked latent graph, and $\hat{y}$ is the final CVS prediction.
}

{
\noindent \textbf{Reconstruction Head Ablations.} We follow the same process as above, but leave the input to the CVS head intact while masking the input to the reconstruction head, thereby optimizing the auxiliary learning objective to maximize domain-invariance in the learned latent graph representations:
\begin{equation}
    G = \Phi_{\text{LG}}(I);\ \mathcal{M}(G, c) = \hat{G}_{\mathcal{R}};\ \hat{y} = \phi_{\text{CVS}}(G);\ \hat{I} = \phi_{\mathcal{R}}(\hat{G}_{\mathcal{R}}).
\end{equation}
Notably, LG-CVS does not include image visual features in the reconstruction input, instead constructing an object-centric feature layout $L_{\text{feat}}$ from $G$ alone before decoding with $\phi_{\text{decoder}}$.
For completeness, we introduce a mechanism to incorporate the image features $H_{\text{backbone}}$ into $L_{\text{feat}}$, which uses predicted object locations to arrange per-object features into a spatial grid, built on simple concatenation:
\begin{equation}
    \hat{L}_{\text{feat}} = [H_{\text{backbone}}; L_{\text{feat}}];\ \hat{I} = \phi_{\text{decoder}}(\hat{L}_{\text{feat}}),
\end{equation}
where we spatially resize $H_{\text{backbone}}$ to match the dimensions of $\hat{L}_{\text{feat}}$.
}
{\subsection{Disentanglement Loss}}
{
While the feature disentanglement ablation studies can shed light on the effect of domain shift on different types of features, ultimately, we are interested in creating a model that uses all available information.
Our key observation is that the masking function $\mathcal{M}$ can also serve as a form of data augmentation during training, and provide robustness to cases where one or more feature categories in $G$ are inaccurate.
To do so, we introduce an auxiliary \textit{disentanglement loss function} $\mathcal{L}_{\text{DIS}}$, that is a linear combination of binary cross entropy loss terms each using the prediction from a different masked graph $\hat{G}$.
Concretely:
\begin{equation}
\fontsize{8pt}{8pt}\selectfont
    \mathcal{L}_{\text{DIS}} = \lambda_{\text{sem}}\mathcal{L}_{\text{CVS}}(\hat{y}_{\text{sem}}, y) + \lambda_{\text{viz}}\mathcal{L}_{\text{CVS}}(\hat{y}_{\text{viz}}, y) + \lambda_{\text{img}}\mathcal{L}_{\text{CVS}}(\hat{y}_{\text{img}}, y),
\end{equation}
where $\hat{y}_{\text{sem}} = \phi_{\text{CVS}}(\hat{G}_{\text{sem}})$, $\hat{y}_{\text{viz}} = \phi_{\text{CVS}}(\hat{G}_{\text{viz}})$, and $\hat{y}_{\text{img}} = \phi_{\text{CVS}}(\hat{G}_{\text{img}})$, $y$ is the ground-truth CVS label, and $\lambda_{\text{sem}}$, $\lambda_{\text{viz}}$, and $\lambda_{\text{img}}$ are loss weighting terms.
}

\begin{table}[h!]
\setlength\tabcolsep{3pt}
\centering
\caption{CVS Classifier Head Inputs Ablation Study in the Domain Generalization Transfer Setting: Endoscapes2023 to Endoscapes-WC70}
\label{tab:cvs_head_ablations}
\begin{tabular}{ccccc}
\multicolumn{3}{c}{\textbf{Feature Type}} & {\textbf{Performance}}                                                                                   & \textbf{}     \\
\begin{tabular}[c]{@{}c@{}}Graph\\ Visual\end{tabular} & \begin{tabular}[c]{@{}c@{}}Graph\\ Semantic\end{tabular} & \begin{tabular}[c]{@{}c@{}}Backbone\\ Image\end{tabular}  & {(mAP)} &               \\ \hline
\cmark                                  & \cmark                                    & \cmark & 27.88 ± 0.37                                                  &  (LG-CVS~\cite{murali2023latent})    \\
\cmark                                  & \cmark                                    & \xmark                                                                                        & 31.37 ± 1.72                                                               \\
\cmark                                  & \xmark                                    & \cmark                                                                                  & 26.31 ± 5.32                                                  & (LG-Visual)   \\
\xmark                                  & \cmark                                    & \xmark                                                                          & 33.81 ± 1.32                                         & (LG-Semantic) \\
\xmark                                  & \cmark                                    & \cmark                                                                                    & \textbf{34.14 ± 0.72}                                                  &               \\
\cmark                                  & \xmark                                    & \xmark                                                                         & 29.36 ± 3.04                                                  &   \\

\end{tabular}
\end{table}

\begin{table}[h!]
\setlength\tabcolsep{3pt}
\centering
\caption{Reconstruction Inputs Ablation Study for \textbf{LG-CVS} in the Domain Generalization Setting, Endoscapes2023 to Endoscapes-WC70.}
\label{tab:reconstruction_ablations}
\begin{tabular}{ccccc}
\multicolumn{3}{c}{\textbf{Feature Type}} & {\textbf{Performance}}                                                                                   & \textbf{}     \\
\begin{tabular}[c]{@{}c@{}}Graph\\ Visual\end{tabular} & \begin{tabular}[c]{@{}c@{}}Graph\\ Semantic\end{tabular} & \begin{tabular}[c]{@{}c@{}}Backbone\\ Image\end{tabular}  & {(mAP)} &               \\ \hline
\cmark & \cmark & \cmark & 27.88 ± 0.37 &   \\
\cmark & \cmark & \xmark & 23.07 ± 2.22 & (LG-CVS~\cite{murali2023latent}) \\
\cmark & \xmark & \cmark & 27.23 ± 0.61  &                \\
\xmark & \cmark & \xmark & \textbf{28.60 ± 2.41} &                 \\
\xmark & \cmark & \cmark & 28.29 ± 2.36      &
\end{tabular}
\end{table}

\subsection{Training}
{As in~\cite{murali2023latent,murali2023endoscapes}, we train all models (except ResNet50) in two-stages, first finetuning a COCO-pretrained Mask-RCNN instance segmentation model on the images with segmentation annotations, then freezing this model and training each object-centric classifier using all images with CVS annotations.
To address class imbalance, we train with an inverse frequency-balanced binary cross entropy loss.
Finally, when training LG-DG, we set $\lambda_{\text{sem}} = 1$, $\lambda_{\text{viz}} = 0.3$, and $\lambda_{\text{img}} = 0.3$.}

\thispagestyle{default}

\section{Experiments and Results}\label{sec2}

As introduced in \hyperref[sec1]{Section 1}, there are traditionally three paradigms for evaluating domain transfer: \textbf{Supervised Domain Adaptation}, \textbf{Unsupervised Domain Adaptation}, and {\textbf{Domain Generalization}}.
Because object-centric models {contain two components: an object detector and a classification model,} we also consider an additional evaluation setting: \textbf{Partially Supervised Domain Adaptation}, where we assume access to CVS labels in the target domain {to train the classification model} but not segmentation labels {to train the object detector}\footnote{This setting represents a very realistic scenario as collecting dense bounding box or segmentation labels is orders of magnitude more expensive than image-level annotations for classification tasks like CVS.}. {Here, we freeze the Mask-RCNN detector trained on the source domain} and train only the second stage of each object-centric method on the target domain.
Finally, we replace the Supervised Domain Adaptation setting with Fully Supervised evaluation, as our primary objective is to establish ceiling performances.
We also omit the Unsupervised Domain Adaptation setting, which warrants a more thorough investigation in future work.

\begin{table*}[h!]
\centering
\caption{Existing Object-Centric Models evaluated in Various Domain Adaptation Settings: Endoscapes2023 to Endoscapes-WC70.}
\label{tab:endo_to_wc}
\begin{tabular}{cccc}
{\textbf{Model}} & \multicolumn{3}{c}{\textbf{Performance (mAP)}}                                      \\
                                & Fully Supervised      & Partially-Supervised                 & Domain Generalization \\ \hline
LG-CVS~\cite{murali2023latent}                         & 33.69 ± 4.60          & 35.63 ± 3.34          & 28.06 ± 3.40              \\
DeepCVS                         & 34.21 ± 2.42          & \textbf{37.83 ± 2.87} & 29.54 ± 1.46     \\
LayoutCVS                       & \textbf{35.43 ± 1.68}          & 35.95 ± 4.11          & \textbf{30.22 ± 1.92}              \\
ResNet50-DetInit               & 27.38 ± 7.03 & 32.34 ± 5.11          & 23.48 ± 3.13              \\
ResNet50                        & 27.58 ± 2.40          & -         & 12.77 ± 2.71       \\
\hline
LG-DG (Ours)                         & \textbf{38.21 ± 2.03}         & \textbf{44.18 ± 2.48}          & \textbf{33.34 ± 2.22}              \\
\end{tabular}
\end{table*}

\begin{table*}[h!]
\centering
\caption{Existing Object-Centric Models evaluated in Various Domain Adaptation Settings: Endoscapes-WC70 to Endoscapes2023.}
\label{tab:wc_to_endo}
\begin{tabular}{cccc}
\textbf{Model}             & \multicolumn{3}{c}{\textbf{Performance (mAP)}}                                     \\
                  & Fully Supervised & Partially-Supervised & Domain Generalization \\ \hline
LG-CVS~\cite{murali2023latent}   & \textbf{64.45 ± 1.30}       & 55.64 ± 0.59                    & 33.83 ± 1.20            \\
DeepCVS           & 58.80 ± 2.06       & 42.84 ± 2.32                    & \textbf{44.30 ± 1.70}            \\
LayoutCVS         & 58.14 ± 0.76       & 43.76 ± 0.70                    & 43.26 ± 1.40            \\
ResNet50-DetInit  & 61.86 ± 2.57       & \textbf{57.63 ± 0.89}                    & 44.24 ± 0.39            \\
ResNet50          & 52.27 ± 3.45                & -                    & 36.11 ± 0.68  \\
\hline

LG-DG (Ours) & \textbf{67.25 ± 0.90}       & \textbf{57.91 ± 2.44}                    & \textbf{47.95 ± 2.40}          \\
\end{tabular}
\end{table*}

\begin{table*}[h!]
\centering
\caption{Per-Class Breakdown of Object Detection Performance (Mask-RCNN, Instance Segmentation mAP) in Target Domain. WC70 refers to Endoscapes-WC70.}
\label{tab:detector}
\begin{tabular}{ccccccccc}
\textbf{Trained On} & \textbf{Tested On} & \textbf{\begin{tabular}[c]{@{}c@{}}Cystic\\ Plate\end{tabular}} & \textbf{\begin{tabular}[c]{@{}c@{}}Calot\\ Triangle\end{tabular}} & \textbf{\begin{tabular}[c]{@{}c@{}}Cystic\\ Artery\end{tabular}} & \textbf{\begin{tabular}[c]{@{}c@{}}Cystic\\ Duct\end{tabular}} & \textbf{Gallbladder} & \textbf{Tool} & \textbf{Avg}  \\ \hline
WC70     & WC70    & 13.1                                                            & 3.0                                                               & 11.9                                                             & 15.2                                                           & \textbf{44.7}        & \textbf{55.2} & \textbf{23.8} \\
Endoscapes2023         & WC70    & \textbf{17.5}                                                   & \textbf{3.9}                                                      & \textbf{14.1}                                                    & \textbf{16.3}                                                  & 39.5                 & 49.1          & 23.4          \\ \hline
WC70 & Endoscapes2023 & 1.8 & 6.9 & 4.4 & 9.4 & 43.4 & 36.9 & 17.13 \\

Endoscapes2023         & Endoscapes2023        & \textbf{9.1}                                                             & \textbf{22.5}                                                              & \textbf{13.4}                                                             & \textbf{19.2}                                                           & \textbf{65.8}                 & \textbf{57.6}          & \textbf{31.3}
\end{tabular}
\end{table*}

{
In \autoref{sec4.1}, we analyze the results of the two latent graph ablation studies: CVS Head and Reconstruction Head (Tables \ref{tab:cvs_head_ablations} and \ref{tab:reconstruction_ablations} respectively).
Then, we move on to the main experiments in \autoref{sec4.2}, where we analyze the domain generalization performance of the four original object-centric methods, a ResNet50 non-object-centric baseline, and our proposed LG-DG (Tables \ref{tab:endo_to_wc} and \ref{tab:wc_to_endo}).
}
Lastly, \autoref{tab:detector} shows the domain generalization performance of each object detector for reference.
{We also show the partially supervised and supervised performances in Tables \ref{tab:endo_to_wc}, \ref{tab:wc_to_endo}, and \ref{tab:detector} to better contextualize our results.}

\label{sec4.1}
\subsection{Latent Graph Ablation Studies}
{
For both ablation studies, we use Endoscapes2023 as the source domain and Endoscapes-WC70 as the target domain.
\autoref{tab:cvs_head_ablations} shows the results of the CVS head ablations (different configurations of $\hat{G}_{\text{CVS}}$); we identify two particular settings of interest:} (1) LG-Semantic, which omits all visual information, and (2) LG-Visual, which omits semantics and uses all the visual information.

We observe that, {in the domain generalization setting}, LG-Semantic attains an mAP of {33.81} mAP, surpassing LG-CVS by {21.2\%}.
{This shows that the various feature modalities do not contribute synergistically when posed with domain shift.
We attribute the particular effectiveness of LG-Semantic to the fact that it only uses the detected objects for downstream CVS prediction, thereby providing robustness to visual domain gap.}
Studying the domain generalization performance of the Mask-RCNN object detector (see Table \ref{tab:detector}) further substantiates this notion: the object detector trained on Endoscapes2023 performs effectively on par with that trained on WC70 when tested on WC70. It even outperforms the latter on the four smaller classes, all of which are critical for CVS assessment.
{Since the object detector generalizes effectively, LG-Semantic performs well.}
Meanwhile, LG-Visual performs markedly worse than LG-Semantic, and slightly worse than LG-CVS, reiterating the notion that semantic features are more robust to domain shifts than visual features.

Table \ref{tab:reconstruction_ablations} shows the results of the {reconstruction head ablations (different configurations of $\hat{G}_{\mathcal{R}}$).
Broadly, we find that including visual features in the reconstruction input, whether graph visual or backbone image features, has a negative impact on domain generalization performance: masking both graph visual features and backbone image features in $G_\mathcal{R}$ yields the highest performance.
One potential explanation for this trend is that, when visual features are part of the reconstruction input, they learn to encode highly domain-specific information like color and texture distributions, degrading the overall domain invariance of the latent graph $G$.
On the other hand, when only semantic information is included, the reconstruction objective only enforces encoding general, perhaps even instance-agnostic, properties of the objects, as the semantic features do not encode properties like color and texture by construction.}

\thispagestyle{default}

\subsection{Main Experiments}
\label{sec4.2}

{
\autoref{tab:endo_to_wc} and \autoref{tab:wc_to_endo} summarize CVS prediction performance for all three domain adaptation settings, tested on Endoscapes2023 and Endoscapes-WC70 respectively.
To recall: Fully Supervised indicates that both the object detector and object-centric classification model are trained and tested on the target domain; Partially Supervised indicates that the object detector is trained on the source domain; Domain Generalization indicates that neither object detector nor object-centric classification model is trained on the target domain.
We show both Fully Supervised and Partially Supervised performance to illustrate the ceiling performance, which we use to contextualize domain generalization performance.}

{
LG-DG consistently outperforms all baseline methods across all settings: for Domain Generalization, LG-DG demonstrates a mean increase of 9.28\% over the best-performing baseline models, LayoutCVS and DeepCVS respectively.
Interestingly, both LayoutCVS and DeepCVS are primarily based on semantic features, with LayoutCVS explicitly ignoring visual information and DeepCVS ineffective at leveraging visual information (shown in~\cite{murali2023latent}); this reinforces our conclusions from the Latent Graph ablation study.
Importantly, LG-DG vastly outperforms LG-CVS: 41.74\% better when tested on Endoscapes2023 and 18.82\% better when tested on Endoscapes-WC70 (average 30.28\%).
This clearly highlights the effectiveness of our proposed optimizations.
Altogether, LG-DG takes great strides in closing the domain gap with no information about the target domain: it is 24.54\% lower than the ceiling performance for Endoscapes-WC70 (Partially Supervised LG-DG) and 28.70\% lower than the ceiling performance for Endoscapes (Fully Supervised LG-DG).}


\thispagestyle{default}

\section{Conclusion}
We investigate the capabilities of object-centric approaches for domain adaptation specifically in the context of fine-grained classification, using Critical View of Safety prediction to measure performance.
We show that object-centric methods {are highly effective for Domain Generalization}, particularly compared to non-object-centric image classifiers.
{Then, leveraging the modularity of object-centric approaches, we propose an optimized method, \textbf{LG-D}omain\textbf{G}en, that includes a novel disentanglement loss to improve robustness to domain shift.}
Future work should seek to {extensively} validate these findings on various surgical procedures and tasks and to {extend these findings} to other settings including UDA and Federated Learning.

\section{Acknowledgments}
This work was supported by French state funds managed by the ANR within the National AI Chair program under Grant ANR-20-CHIA-0029-01 (Chair AI4ORSafety). This work was granted access to the HPC resources of IDRIS under the allocation AD011013523R1 made by GENCI.

\noindent{\bf Ethical Approval:} This article does not contain any studies with human participants or animals performed by any of the authors.

\noindent{\bf Competing Interests:} The authors declare no conflict of interest.

\noindent{\bf Informed Consent:} This manuscript does not contain any patient data.

\noindent{\bf Code Availability:} The source code is publicly available at \url{https://github.com/CAMMA-public/SurgLatentGraph}.

\bibliographystyle{sn-basic}
\bibliography{main}

\begin{thebibliography}{22}
\providecommand{\natexlab}[1]{#1}
\providecommand{\url}[1]{{#1}}
\providecommand{\urlprefix}{URL }
\providecommand{\doi}[1]{\url{https://doi.org/#1}}
\providecommand{\eprint}[2][]{\url{#2}}

\bibitem[{Basak and Yin(2023)}]{basak2023semi}
Basak H, Yin Z (2023) Semi-supervised domain adaptive medical image segmentation through consistency regularized disentangled contrastive learning. In: MICCAI, Springer, pp 260--270

\bibitem[{Chen et~al(2023)Chen, Pan, Ye, Cui, and Xia}]{trid}
Chen Z, Pan Y, Ye Y, et~al (2023) Treasure in distribution: A domain randomization based multi-source domain generalization for 2d medical image segmentation. In: MICCAI. Springer Nature Switzerland, Cham, pp 89--99

\bibitem[{Choi et~al(2021)Choi, Jung, Yun, Kim, Kim, and Choo}]{choi2021robustnet}
Choi S, Jung S, Yun H, et~al (2021) Robustnet: Improving domain generalization in urban-scene segmentation via instance selective whitening. In: CVPR, pp 11580--11590

\bibitem[{Grammatikopoulou et~al(2021)Grammatikopoulou, Flouty, Kadkhodamohammadi, Quellec, Chow, Nehme, Luengo, and Stoyanov}]{grammatikopoulou2021cadis}
Grammatikopoulou M, Flouty E, Kadkhodamohammadi A, et~al (2021) Cadis: Cataract dataset for surgical rgb-image segmentation. Medical Image Analysis 71

\bibitem[{Hamoud et~al(2023)Hamoud, Jamal, Srivastav, Mutter, Padoy, and Mohareri}]{hamoud2023st}
Hamoud I, Jamal MA, Srivastav V, et~al (2023) St(or)$^2$: Spatio-temporal object level reasoning for activity recognition in the operating room. In: Medical Imaging with Deep Learning

\bibitem[{Hao et~al(2023)Hao, Hu, Lin, Wang, Li, Fu, Duan, and Liu}]{hao2023act}
Hao L, Hu Y, Lin W, et~al (2023) Act-net: Anchor-context action detection in surgery videos. In: MICCAI, Springer, pp 196--206

\bibitem[{Holm et~al(2023)Holm, Ghazaei, Czempiel, {\"O}zsoy, Saur, and Navab}]{holm2023dynamic}
Holm F, Ghazaei G, Czempiel T, et~al (2023) Dynamic scene graph representation for surgical video. In: Proceedings of the IEEE/CVF International Conference on Computer Vision, pp 81--87

\bibitem[{Kassem et~al(2022)Kassem, Alapatt, Mascagni, AI4SafeChole, Karargyris, and Padoy}]{kassem2022federated}
Kassem H, Alapatt D, Mascagni P, et~al (2022) Federated cycling (fedcy): Semi-supervised federated learning of surgical phases. IEEE Transactions on Medical Imaging

\bibitem[{Mascagni et~al(2021)Mascagni, Vardazaryan, Alapatt, Urade, Emre, Fiorillo, Pessaux, Mutter, Marescaux, Costamagna et~al}]{mascagni2021artificial}
Mascagni P, Vardazaryan A, Alapatt D, et~al (2021) Artificial intelligence for surgical safety: automatic assessment of the critical view of safety in laparoscopic cholecystectomy using deep learning. Annals of Surgery

\bibitem[{Mottaghi et~al(2022)Mottaghi, Sharghi, Yeung, and Mohareri}]{mottaghi2022adaptation}
Mottaghi A, Sharghi A, Yeung S, et~al (2022) Adaptation of surgical activity recognition models across operating rooms. In: MICCAI, Springer, pp 530--540

\bibitem[{Murali et~al(2023{\natexlab{a}})Murali, Alapatt, Mascagni, Vardazaryan, Garcia, Okamoto, Costamagna, Mutter, Marescaux, Dallemagne et~al}]{murali2023endoscapes}
Murali A, Alapatt D, Mascagni P, et~al (2023{\natexlab{a}}) The endoscapes dataset for surgical scene segmentation, object detection, and critical view of safety assessment: Official splits and benchmark. arXiv preprint arXiv:231212429

\bibitem[{Murali et~al(2023{\natexlab{b}})Murali, Alapatt, Mascagni, Vardazaryan, Garcia, Okamoto, Mutter, and Padoy}]{murali2023encoding}
Murali A, Alapatt D, Mascagni P, et~al (2023{\natexlab{b}}) Encoding surgical videos as latent spatiotemporal graphs for object and anatomy-driven reasoning. In: MICCAI, Springer, pp 647--657

\bibitem[{Murali et~al(2023{\natexlab{c}})Murali, Alapatt, Mascagni, Vardazaryan, Garcia, Okamoto, Mutter, and Padoy}]{murali2023latent}
Murali A, Alapatt D, Mascagni P, et~al (2023{\natexlab{c}}) Latent graph representations for critical view of safety assessment. IEEE Transactions on Medical Imaging pp 1--1

\bibitem[{{\"O}zsoy et~al(2023){\"O}zsoy, Czempiel, Holm, Pellegrini, and Navab}]{ozsoy2023labrad}
{\"O}zsoy E, Czempiel T, Holm F, et~al (2023) Labrad-or: Lightweight memory scene graphs for accurate bimodal reasoning in dynamic operating rooms. arXiv preprint arXiv:230313293

\bibitem[{Pang et~al(2022)Pang, Islam, Mitheran, Seenivasan, Xu, and Ren}]{pang2022rethinking}
Pang W, Islam M, Mitheran S, et~al (2022) Rethinking feature extraction: Gradient-based localized feature extraction for end-to-end surgical downstream tasks. IEEE Robotics and Automation Letters 7(4):12623--12630

\bibitem[{Sestini et~al(2023)Sestini, Rosa, De~Momi, Ferrigno, and Padoy}]{sestini2023fun}
Sestini L, Rosa B, De~Momi E, et~al (2023) Fun-sis: A fully unsupervised approach for surgical instrument segmentation. Medical Image Analysis 85:102751

\bibitem[{Sharma et~al(2023)Sharma, Nwoye, Mutter, and Padoy}]{sharma2023surgical}
Sharma S, Nwoye CI, Mutter D, et~al (2023) Surgical action triplet detection by mixed supervised learning of instrument-tissue interactions. In: MICCAI, Springer, pp 505--514

\bibitem[{Sohan and Basalamah(2023)}]{sohan2023systematic}
Sohan MF, Basalamah A (2023) A systematic review on federated learning in medical image analysis. IEEE Access

\bibitem[{Srivastav et~al(2022)Srivastav, Gangi, and Padoy}]{srivastav2021adaptor}
Srivastav V, Gangi A, Padoy N (2022) Unsupervised domain adaptation for clinician pose estimation and instance segmentation in the operating room. In: Medical Image Analysis

\bibitem[{Twinanda et~al(2016)Twinanda, Shehata, Mutter, Marescaux, De~Mathelin, and Padoy}]{twinanda2016endonet}
Twinanda AP, Shehata S, Mutter D, et~al (2016) Endonet: a deep architecture for recognition tasks on laparoscopic videos. IEEE transactions on medical imaging 36(1):86--97

\bibitem[{Wang et~al(2019)Wang, Bu, and Breckon}]{wang2019unifying}
Wang Q, Bu P, Breckon TP (2019) Unifying unsupervised domain adaptation and zero-shot visual recognition. In: 2019 International Joint Conference on Neural Networks (IJCNN), IEEE, pp 1--8

\bibitem[{Xu et~al(2022)Xu, Zhang, Yu, Zhao, Bian, Liu, Wang, Ge, and Qian}]{xu2022deep}
Xu J, Zhang Q, Yu Y, et~al (2022) Deep reconstruction-recoding network for unsupervised domain adaptation and multi-center generalization in colonoscopy polyp detection. Computer Methods and Programs in Biomedicine 214:106576

\end{thebibliography}
\end{document}